%% file: 00_Main.tex
\documentclass[11pt,letterpaper]{article}

\usepackage{xcolor}
\definecolor{navy}{rgb}{0,0.1,0.4}
\definecolor{navy-cap}{rgb}{0,0.1,0.5}
\definecolor{navy-ref}{rgb}{0,0.3,1}
\definecolor{lgray}{gray}{0.90}
\definecolor{dkgreen}{rgb}{0,0.6,0}
\definecolor{gray}{rgb}{0.5,0.5,0.5}
\definecolor{mauve}{rgb}{0.58,0,0.82}

\usepackage{sectsty}
\sectionfont{\fontsize{12}{15}\selectfont \color{navy}}
\subsectionfont{\fontsize{12}{15}\selectfont \color{navy}}
\subsubsectionfont{\fontsize{12}{15}\selectfont \color{navy}}

\usepackage[margin = 1 in]{geometry}

\usepackage{times}
\usepackage[font={color=navy-cap, small}, labelfont={bf}, labelsep=space, justification=justified, skip=0pt]{caption}
\usepackage[bottom, hang,flushmargin]{footmisc} 

\usepackage{titlesec}
\usepackage{tikz}
\titlespacing{\section}{0pt}{0.75em}{0.25em}
\titlespacing{\subsection}{0pt}{0.5em}{0.25em}
\titlespacing{\subsubsection}{0pt}{0.5em}{0.25em}

\usepackage{array,longtable}
\usepackage{xltabular}
\usepackage{multirow}
\usepackage{pifont}
\usepackage[font={color=navy-cap, footnotesize}, labelfont={bf}, labelsep=space, justification=justified, skip=3pt]{caption}
\usepackage{subcaption}
\usepackage{booktabs} 
\usepackage{graphicx}
\usepackage{wrapfig}
\usepackage{enumitem}
\usepackage{hyperref}
\hypersetup{colorlinks = true, linkcolor = navy-ref}
\usepackage{textcomp}
\usepackage{url}
\usepackage{amssymb}
\usepackage{bm}
\usepackage{amsmath}

\usepackage{dirtytalk}
\usepackage{pdfpages} 
\usepackage{soul} 
\usepackage{fancyhdr}
\usetikzlibrary{patterns}
\usepackage{pgfplots}
\usepackage{floatrow}
\usepackage{setspace}
\usepackage{bbm}

\usepackage[
backend=biber,
style=numeric-comp,
sorting=none,
maxnames = 10
]{biblatex}
\usepackage[nameinlink]{cleveref} 

\usepackage{colortbl}
\usepackage{hhline}

\newcommand{\alloy}{17-4 PH SS}
\newcommand{\secalloy}{316L SS}

\newcommand{\porosity}{$\phi$}
\newcommand{\hardness}{$HV$}

\newcommand{\betab}{\boldsymbol{\beta}}

\newcommand{\thetab}{\boldsymbol{\theta}}

\newcommand{\omegab}{\boldsymbol{\omega}}

\newcommand\braces[1]{\mathopen{}\left\{#1\right\}\mathclose{}}
\newcommand{\xsspace}{\mathbb{X}}
\newcommand{\rsspace}{\mathbb{R}}

\DeclareMathOperator*{\E}{\mathbb{E}}

\newcommand{\yb}{\boldsymbol{y}}

\newcommand{\xb}{\boldsymbol{x}}
\newcommand{\Xb}{\boldsymbol{X}}

\newcommand{\mb}{\boldsymbol{m}}

\newcommand{\Cb}{\boldsymbol{C}}

\usepackage[ruled,vlined]{algorithm2e}

\usepackage{bbm}
\usepackage{xfrac}
\usepackage{enumitem}

\usepackage{amsthm}

\usepackage{enumitem}
\usepackage{cancel}
\usepackage{mathtools}

\usepackage{placeins}

\usepackage{multicol}

    \makeatletter
\def\@fnsymbol#1{\ensuremath{\ifcase#1\or \dagger\or *\or \ddagger\or
   \mathsection\or \mathparagraph\or \|\or **\or \dagger\dagger
   \or \ddagger\ddagger \else\@ctrerr\fi}}
    \makeatother

\title{A preliminary data fusion study to assess the feasibility of Foundation Process-Property Models in Laser Powder Bed Fusion}
\date{\vspace{-5ex}}
\usepackage{authblk}
\author[1]{Oriol Vendrell-Gallart}
\author[1]{Nima Negarandeh}
\author[1]{Zahra Zanjani Foumani}
\author[2]{Mahsa Amiri}
\author[2,3]{Lorenzo Valdevit\thanks{Corresponding Authors: valdevit@uci.edu, raminb@uci.edu}}
\author[1,4]{Ramin Bostanabad$^\dagger$}

\small\affil[1]{Department of Mechanical and Aerospace Engineering, University of California, Irvine, CA, 92697, USA}
\affil[2]{Materials and Manufacturing Technology Program, University of California, Irvine, CA, 92697, USA}
\affil[3]{Department of Materials Science and Engineering, University of California, Irvine, CA, 92697, USA}
\affil[4]{Department of Civil and Environmental Engineering, University of California, Irvine, CA, 92697, USA}
\begin{document}
    \pagenumbering{arabic}
    \maketitle
    \sloppy 
    \input{11_Abstract}
    \input{12_Introduction}
    \input{13_Methodology}
    \input{14_Results}

    \input{15_Conclusion}
    \printbibliography
\end{document}

%% file: 11_Abstract.tex
\noindent \textcolor{navy}{\textbf{Abstract}}

Foundation models are at the forefront of an increasing number of critical applications. In regards to technologies such as additive manufacturing (AM), these models have the potential to dramatically accelerate process optimization and, in turn, design of next generation materials. A major challenge that impedes the construction of foundation process-property models is data scarcity. To understand the impact of this challenge, and since foundation models rely on data fusion, in this work we conduct controlled experiments where we focus on the transferability of information across different material systems and properties. More specifically, we generate experimental datasets from 17-4 PH and 316L stainless steels (SSs) in Laser Powder Bed Fusion (LPBF) where we measure the effect of five process parameters on porosity and hardness. 
We then leverage Gaussian processes (GPs) for process-property modeling in various configurations to test if knowledge about one material system or property can be leveraged to build more accurate machine learning models for other material systems or properties. 
Through extensive cross-validation studies and probing the GPs' interpretable hyperparameters, 
we study the intricate relation among data size and dimensionality, complexity of the process-property relations, noise, and characteristics of machine learning models. 
Our findings highlight the need for structured learning approaches that incorporate domain knowledge in building foundation process-property models rather than relying on uninformed data fusion in data-limited applications.

\noindent \textcolor{navy}{\textbf{Keywords:}} 
Foundation Models; Transfer Learning; Data Fusion; Additive manufacturing; Gaussian process.

%% file: 12_Introduction.tex
\section{Introduction}  \label{sec: intro}

The increasing need for lightweight, high-performance materials with complex geometries has driven the advancement of additive manufacturing (AM) technologies \cite{DebRoy2018AdditiveProperties}. Among the different AM processes, laser powder bed fusion (LPBF) is a leading technique for producing metallic components that offer outstanding mechanical properties and design versatility \cite{Fields2024InvestigationProperties}.

The properties of LPBF-built parts are heavily influenced by various process parameters such as laser power and speed, scan direction, and deposition layer thickness. Fully understanding the complex relationships between process parameters, microstructural evolution, and mechanical properties in LPBF, with sufficient accuracy for effective optimization, remains a significant challenge \cite{Fields2024MicrostructuralFusion}. The vast parameter space inherent in LPBF makes brute-force experimentation impractical. As a result, predictive modeling of
process-property relations in LPBF has traditionally relied on domain knowledge and trial-and-error methods \cite{Agrawal2022High-throughputMaterials}. Therefore, the development of robust, generalizable models capable of accurately predicting material properties from process parameters is crucial for optimizing AM processes and accelerating material design. This motivates the exploration of foundation process-property models to accelerate the optimization and design of materials with desired properties.

In recent years, foundation models \cite{bommasani2022opportunitiesrisksfoundationmodels} —large-scale machine learning (ML) models capable of generalizing across diverse datasets— have shown remarkable success in domains such as natural language processing and computer vision where vast amounts of data are readily available. However, in materials science and manufacturing processes such as LPBF, the availability of large, high-quality datasets is a significant challenge. This raises the issue of whether foundation models can be effectively leveraged for material property prediction despite limited data availability.

To address this question, in this work we focus on a controlled environment where we evaluate the transferability of knowledge between two distinct stainless steel (SS) material systems, namely, \alloy~and \secalloy. Our underlying hypothesis is that these materials contain transferable process-property information, i.e., knowledge from one system can be leveraged to build more accurate process-property ML models for the other material system. However, this assumption remains largely untested in the literature, particularly with respect to LPBF.
To test our hypothesis, we model process-property relations via Gaussian processes (GPs) whose training data is based on either one or both steel types. 

Our studies involve data fusion which we formulate via GPs because they not only are extremely effective in learning from small data, but also have explainable parameters which supports decision making. 
Our findings have significant implications for materials design and process optimization and they highlight the need for \textit{principled} transfer learning for building foundation process-property models where domain knowledge, such as physics-informed priors or hierarchical modeling \cite{amiri2024unveiling}, is leveraged for data fusion. 

The rest of the paper is organized as follows. We introduce our material characterization and ML approaches in Section \ref{sec: method} and then assess the feasibility of developing foundation process-property models in Section \ref{sec: results}. We conclude the paper in Section \ref{sec: conclusion} by summarizing our contributions and providing future research directions.  

%% file: 13_Methodology.tex
\section{Methods} \label{sec: method}
We first elaborate on the experimental data generation process in Section \ref{sec: AM_design} and then in Sections \ref{sec: GPs_intro} and \ref{sec: MTGPs_intro} provide some background on GPs which serve as the workhorse of our studies. Finally, our approach is explained in Section \ref{sec: ML_method} where we assess the potential advantages of learning two mechanical properties (namely, porosity and hardness) of \alloy~and \secalloy~under various settings.

\subsection{Experimental Data Collection} \label{sec: AM_design}
The \alloy~data used in this study is derived  from \cite{amiri2024unveiling} which consists of $270$ process parameter combinations generated using a Sobol sequence. These combinations include variations in laser power ($p$) ranging from $80$ to $400$ W, laser scan speed ($v$) from $150$ to $1500$ mm/s, powder layer thickness ($l$) from $20$ to $75$ $\mu$m, hatch spacing ($h$) from $70$ to $120$ $\mu$m, and two scan rotation angles ($sr$) of $67$ or $90$ degrees. These parameters were used to fabricate $270$ cube-shaped samples ($2 \times 2 \times 2$ mm), numbered $1$ to $270$, using the HT setting outlined in \cite{amiri2024unveiling}. Detailed information on the HT setting and the $270$ combinations can be found in \cite{amiri2024unveiling}. 

To study the feasibility of information transfer from \alloy~to \secalloy~and vice versa, an additional $270$ cube-shaped \secalloy~samples are fabricated for this study. These samples are built using the same process conditions and manufacturing approach as \alloy, detailed in \cite{amiri2024unveiling}. 

\begin{figure}[!t]
    \centering
    \begin{subfigure}[b]{0.25\textwidth}
        \includegraphics[width=\textwidth]{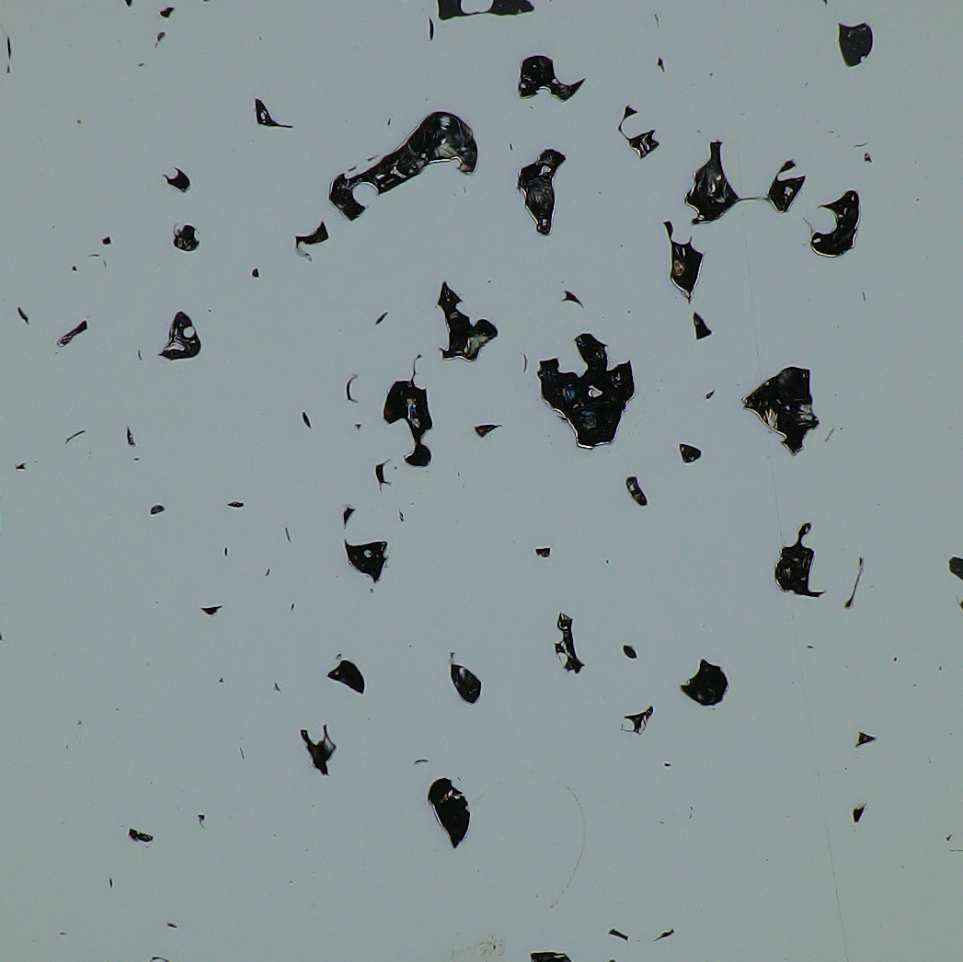}
        \caption{Microstructural image}
        \label{fig:image1}
    \end{subfigure}
    \hspace{0.01\textwidth}
    \begin{subfigure}[b]{0.25\textwidth}
        \includegraphics[width=\textwidth]{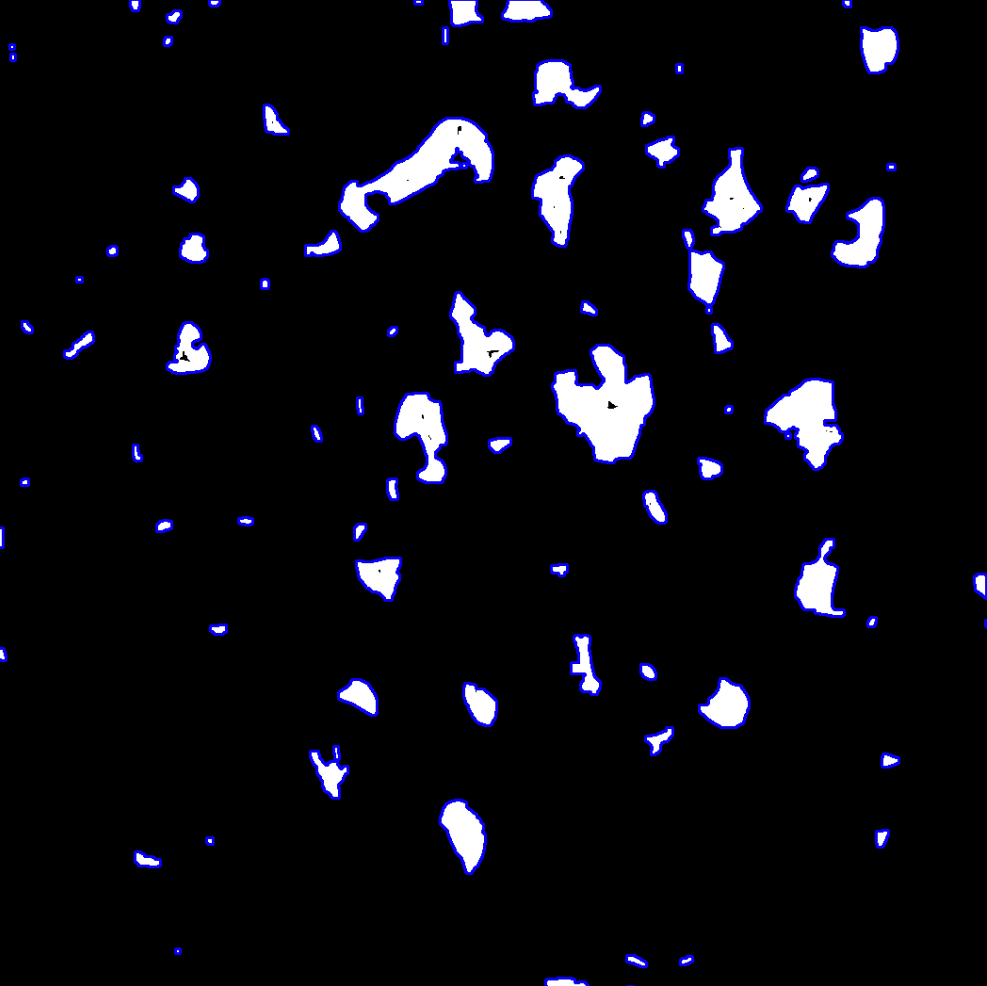}
        \caption{Pore contours}
        \label{fig:image2}
    \end{subfigure}
    \caption{Pores extracted from the microstructural image of a cuboid sample made out of \alloy.}
    \label{fig:pores}
\end{figure}

We analyze the microstructural images of the $270$ cuboids from each material to extract porosity features. In our studies we consider two cases: extracting the average porosity (\porosity) using simple thresholding or conducting image analysis to extract a set of features that describe the spatial distribution and geometry of pores (e.g. statistics of pore radius, perimeter, location, etc.). For the latter case, the images are first cropped to eliminate the borders and then binarized via Otsu's thresholding method \cite{4310076}. Morphological dilation \cite{4767941} is applied to ensure distinct separation of pores, followed by the watershed algorithm \cite{87344} to segment individual pores, see Figure \ref{fig:pores}. Connected component analysis \cite{10.1145/321356.321357} labels each pore region, and the area of each pore is calculated by counting its pixels. The total porosity (\porosity) is determined as the ratio of the combined pore area to the total image area, expressed as a percentage. 
For brevity, in Section \ref{sec: results} we present the results for the former case and suffice to only discussing the outcomes of our studies for the latter case.

\begin{figure*}[!b]
    \centering
    \begin{subfigure}[b]{0.45\textwidth}
        \includegraphics[width=\textwidth]{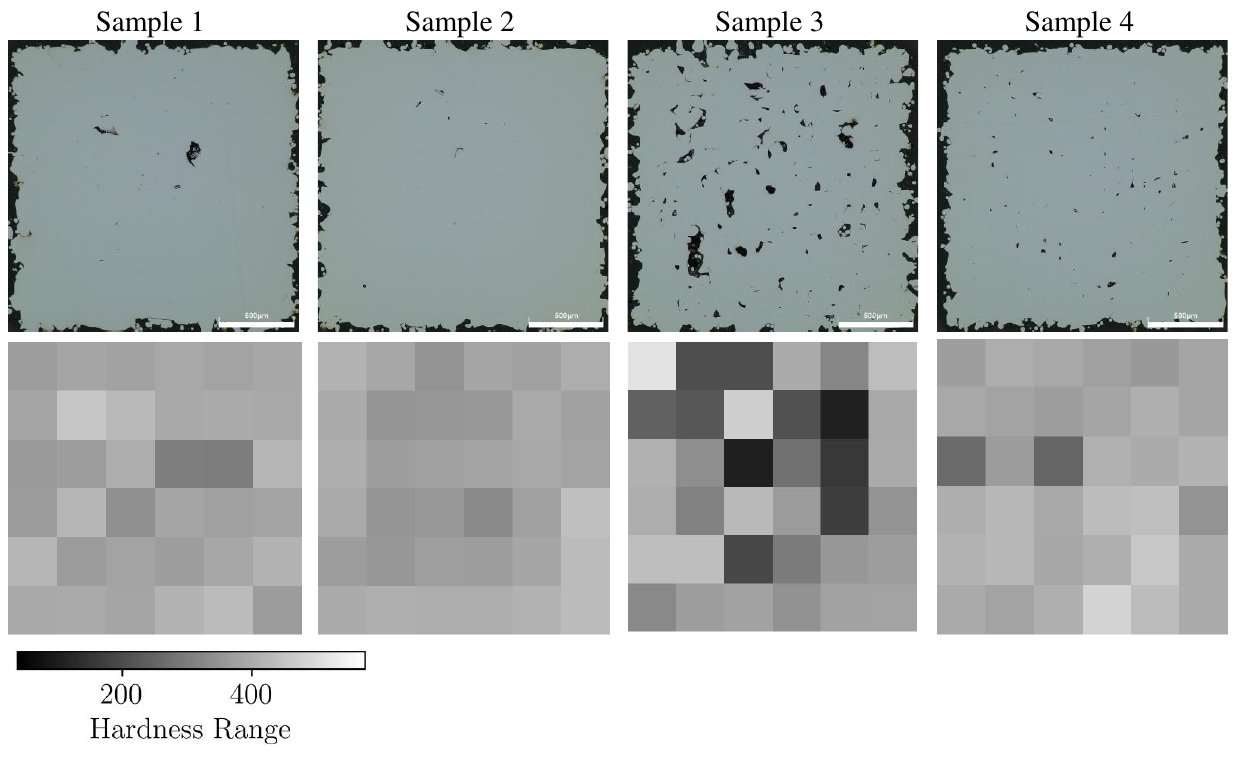}
        \caption{\alloy}
        \label{fig: hardness_174}
    \end{subfigure}
    \begin{subfigure}[b]{0.45\textwidth}
        \includegraphics[width=\textwidth]{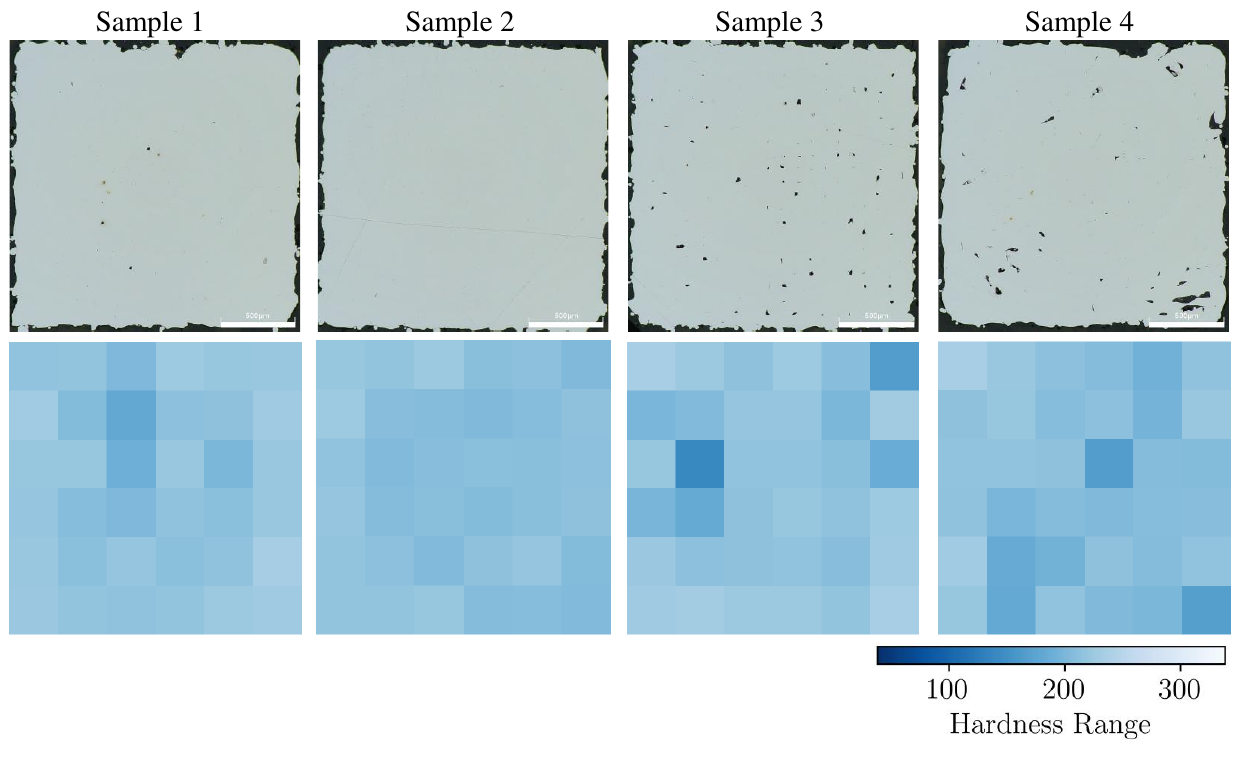}
        \caption{\secalloy}
        \label{fig: hardness_316}
    \end{subfigure}
    \caption{Hardness maps and optical microscopy images of representative samples for (a) \alloy~and (b) \secalloy. We use distinct colors for the hardness maps to aid with visualization and comparison of the hardness range across the two material systems.}
    \label{fig: hardness_maps}
\end{figure*}
In addition to the microstructural analysis, we conducted Vickers microindentation hardness measurements on a $6\times6$ grid on each cuboid to obtain its hardness map, see Figure \ref{fig: hardness_maps}. We take the median hardness value (\hardness) in each hardness map to reduce not only the dimensionality, but also the sensitivity to outliers since pores significantly affect the indentation results.
\subsection{Gaussian Processes} \label{sec: GPs_intro}
GPs \cite{10.5555/1162254} are probabilistic models that assume the training data follows a multivariate normal distribution, where the mean vector and covariance matrix are functions of the inputs.  
Consider a training dataset \(\braces{\xb^{(i)}, y^{(i)}}_{i=1}^n\), where \(\xb = [x_1, ..., x_{dx}] \in \xsspace \subset \rsspace^{dx}\) represents the input variables, and \(y^{(i)} = y(\xb^{(i)}) \in \rsspace\) denotes the corresponding response\footnote{Superscripts enclosed in parentheses enumerate samples. For example, \(x^{(i)}\) refers to the \(i^{th}\) sample in the dataset, whereas \(x_i\) represents the \(i^{th}\) component of the vector \(\xb = [x_1, \cdots, x_{dx}]\).}. Given \(\yb = [y^{(1)}, \cdots, y^{(n)}]^T\) and \(\Xb\), where the \(i^{th}\) row corresponds to \(\xb^{(i)}\), our objective is to predict \(y(\xb^*)\) at an arbitrary point \(\xb^* \in \xsspace\).  

Under this framework, we assume that \(\yb\) is a realization of a GP with the following parametric mean and covariance functions:
\begin{subequations} 
    \begin{equation} 
        \mathbb{E}[y(\xb)] = m(\xb; \betab),
        \label{eq: gp-mean}
    \end{equation}
    \begin{equation} 
        \text{cov}\left(y(\xb), y(\xb')\right) = c(\xb, \xb'; \sigma^2, \thetab) = \sigma^2 r(\xb, \xb'; \thetab)
        \label{eq: gp-cov}
    \end{equation}
    \label{eq: gp-mean and cov}
\end{subequations} 
\noindent where $\betab$, $\sigma^2$, and $\thetab$ are the parameters of the mean and covariance functions. The mean function in GPs can take various forms, from simple polynomials to intricate structures like feed-forward neural networks (FFNN). However, many GP applications often use a constant mean function \(m(\mathbf{x}; \boldsymbol{\beta}) = \beta\) which suggests that the predictive power of the GP mainly depends on its kernel function.

In the covariance function, \(\sigma^2\) stands for the process variance, and \(r(\cdot, \cdot)\) is the correlation function parametrized by $\thetab$. Popular choices for \(r(\cdot, \cdot)\) include the Gaussian, power exponential, and Matérn correlation functions. In our approach, we utilize the Gaussian kernel:
\begin{equation} 
    r\left(\xb, \xb^{\prime}; \omegab \right) = 
            \exp \left\{-\sum_{i=1}^{dx} 10^{\omega_i}(x_i-x_i^{\prime})^2 \right \}  
    \label{eq: rbf-kernel}
\end{equation}
\noindent where $\omega=\thetab$.
The inductive bias that the kernels encode into the learning process is that close-by input vectors $\xb$~and $\xb^{\prime}$ have similar (i.e., correlated) output values. The degree of this correlation is quantified by the interpretable length-scale (aka roughness) parameters, where the magnitude of \(10^{\omega_i}\) is directly related to the response fluctuations along \(x_i\).

To estimate the parameters of a GP, we leverage maximum likelihood estimation (MLE) which is equivalent to minimizing the negative of the marginal likelihood:
\begin{equation} 
    \begin{split}
        [\widehat{\betab}, \widehat{\sigma}^2, \widehat{\thetab}] 
        &= \underset{\betab, \sigma^2, \thetab}{\operatorname{argmin}} \hspace{2mm} 
        \frac{1}{2} \log (|\Cb|) 
        + \frac{1}{2} (\yb - \mb)^T \Cb^{-1} (\yb - \mb)
    \end{split}
    \label{eq: map-gp}
\end{equation}
\noindent where $\Cb$ is the covariance matrix with entries $C_{ij} = c(\xb^{(i)}, \xb^{(j)}; \sigma^2, \thetab)$ and $\mb$ is an $n \times 1$ vector whose $i^{th}$ element is $m_i=m(\xb^{(i)}; \betab)$. Once the parameters are estimated, predictions at a new point $\xb^*$ are obtained using:
\begin{subequations} 
    \begin{equation} 
        \E[y(\xb^*)] = 
        m(\xb^*; \widehat{\betab}) + c(\xb^*, \Xb; \widehat{\thetab}, \widehat{\sigma}^2) \boldsymbol{C}^{-1}(\yb-\mb)
        \label{eq: gp-mean-scalar}
    \end{equation}
    \begin{equation}
    \begin{split}
        \text{cov}(y(\xb^*), y(\xb^*)) &= c(\xb^*, \xb^*; \widehat{\thetab}, \widehat{\sigma}^2) - c(\xb^*, \Xb; \widehat{\thetab}, \widehat{\sigma}^2) \Cb^{-1} 
        c(\Xb, \xb^*; \widehat{\thetab}, \widehat{\sigma}^2)
    \end{split}
    \label{eq: gp-var-scalar}
\end{equation}
    \label{eq: gp-mean-var-scalar}
\end{subequations}    
\noindent where  $c(\xb^*, \Xb; \widehat{\thetab}, \widehat{\sigma}^2)$ represents the covariance between $\xb^*$ and the training points.

\subsection{Multi-Task Gaussian Processes (MTGPs)}\label{sec: MTGPs_intro}
MTGPs' \cite{NIPS2007_66368270} goal is to model multiple correlated tasks (outputs) simultaneously. Let $\yb(\xb) = [y_1(\xb), \dots, y_G(\xb)]$ represent $G$ tasks where each task $y_g(\xb)$ is a GP.  Then, the joint covariance function of a MTGP can be defined as:
\begin{equation}
    \text{cov}_{\text{MT}}(y_g(\xb), y_{g'}(\xb^{\prime})) = c(\xb, \xb'; \sigma^2, \thetab) c_{gg'}
\end{equation}
\noindent where $c(\xb, \xb'; \sigma^2, \thetab)$ is the base kernel (e.g., RBF) and $c_{gg'}$ is a look-up table encoding correlations between $g$ and $g'$. Thus, the covariance matrix of the training data is $\Cb_{MT} = \mathbf{C} \otimes \boldsymbol{C}_T$, where $\otimes$ denotes Kronecker product \cite{van2000ubiquitous} and $\boldsymbol{C}_T \in \mathbb{R}^{G \times G}$ is a positive semi-definite matrix learned during training.


Given observations \(\{\xb^{(i)}, \yb_g^{(i)} \}_{i=1}^{n}\) with $g= 1,\dots,G$, the posterior predictive distribution for a new input $\xb^*$ is defined by the following mean vector and covariance matrix:
\begin{subequations} 
    \begin{equation} 
        \E[\yb(\xb^*)] = 
        m(\xb^*; \widehat{\betab}) + (c(\xb^*, \Xb; \widehat{\thetab}, \widehat{\sigma}^2) \otimes \boldsymbol{C}_T) \boldsymbol{C}_{MT}^{-1}(\yb-\mb)
        \label{eq: gp-mean-scalar}
    \end{equation}
    \begin{equation}
    \begin{split}
        &\text{cov}(\yb(\xb^*), \yb(\xb^*)) = c(\xb^*, \xb^*; \widehat{\thetab}, \widehat{\sigma}^2) \otimes \boldsymbol{C}_T \\
        &\quad - \big(c(\xb^*, \Xb; \widehat{\thetab}, \widehat{\sigma}^2) \otimes \boldsymbol{C}_T\big)
        \Cb_{MT}^{-1} 
        \big(c(\Xb, \xb^*; \widehat{\thetab}, \widehat{\sigma}^2) \otimes \boldsymbol{C}_T\big)
    \end{split}
    \label{eq:gp-var-scalar}
\end{equation}
    \label{eq: gp-mean-var-scalar}
\end{subequations}    
\noindent where $\yb^{(i)}=\{y_1^{(i)},\dots,y_g^{(i)}\}, \forall g=1,\dots,G$, and $\mb$ is an $(n \times G) \times 1$ vector of mean values.

\subsection{Data Fusion for Process-Property Modeling} \label{sec: ML_method}
Central to the development of foundation process-property models are two key questions: (1) whether information can be transferred from one material to another, and (2) whether information can be transferred between material properties by leveraging their correlations, given a shared input feature space. Addressing these questions is critical for enabling generalization across diverse material systems and properties, ultimately accelerating materials design and process optimization. 

To analyze these aspects via small datasets, we argue that an explainable framework is essential — one that not only provides accurate predictions but also offers insights into the underlying structure of the data. GPs are particularly well-suited for this purpose due to their inherent explainability, which allows for the interpretation of relationships between inputs and outputs, as well as the quantification of uncertainties. In this study, we employ GPs in two configurations, namely, single-output GPs (SOGPs) and MTGPs, both with and without data fusion to systematically explore these questions and uncover the mechanisms governing transferability and correlations in material systems.

In our studies, we build SOGPs independently for each property—\hardness~and \porosity—under two scenarios: $(1)$ using data from a single material (\alloy~ or \secalloy) and $(2)$ using fused data from both materials. These models serve as baselines to evaluate the impact of data fusion on predicting a single material property. 
The assumption we aim to test via these studies is that fusing materials data increases the generalizability of ML models. The physical justification behind this assumption is that \alloy~and \secalloy~alloys share some correlations in terms of how their mechanical properties are affected by LPBF process parameters. 

In addition to SOGPs, we use MTGPs to investigate the potential benefits of property correlation. Specifically, we use MTGPs to jointly learn \porosity~and \hardness~using either data from one material or fused data from two alloys.


For combining the two materials data, we follow \cite{eweis2022data} where data fusion is converted to a latent space learning problem. This conversion is achieved by augmenting each input with an additional categorical variable, \( s \), which represents the source of the input—in our case, the material type, specifically  $s \in \{'\text{\alloy}', '\text{\secalloy}'\}$. Subsequently, we concatenate the training data from all sources to build a fused model. Once trained, predicting the objective function value at \( \xb^{*} \) for material \( j \) involves concatenating the query point with the corresponding categorical variable \( s \) and feeding the combined input into the fused model.

To effectively handle the added categorical feature with GPs, we modify the method proposed in \cite{oune2021latent} where high-dimensional quantitative priors are first assigned to categorical variables and then these priors are passed through a parametric embedding function to generate a low-dimensional latent representation.  
Since we only have two types of material in this paper, our assigned priors are only one-dimensional (e.g., one-hot encoding) which reduces the parameter number of the embedding function to one. While maintaining explainability, this choice dramatically reduces the number of hyperparameters and, in turn, reduces potential overfitting issues. 

To ensure robust evaluation, we use $5$-fold cross-validation (CV) to calculate root mean squared error (RMSE):
\begin{equation}
    \text{RMSE} = \sqrt{\frac{1}{n_{test}} \sum_{i=1}^{n_{test}} (y^{(i)} - \hat{y}^{(i)})^2}
\end{equation}
\noindent where $y^{(i)}=y({\xb^*}^{(i)})$ and $\hat{y}^{(i)}=\E[y({\xb^*}^{(i)})]$ represents the output and predicted value for test sample ${\xb^*}$, respectively, and $n_{test}$ is the number of test samples.

By comparing the performance of SOGPs and MTGPs, with and without data fusion, we aim to test two hypotheses: (1) combining data from \alloy~and \secalloy~improves predictive accuracy by leveraging some underlying shared relation in the process-property links, and (2) jointly modeling \porosity~and \hardness~enhances predictive accuracy of both properties by capturing correlations between them. This systematic evaluation provides insights into the feasibility of transfer learning and data fusion in additive manufacturing, particularly in data-limited scenarios, and highlights the role that explainable models play in building robust foundation process-property models.

%% file: 14_Results.tex
\section{Results and Discussions} \label{sec: results}
We begin by exploring the relationships between process parameters and material properties in Section \ref{sec:data-correlation}, highlighting the stochasticity of the data and the potential for process-property modeling. Next, we compare the performance of different model configurations in Section \ref{sec:models-comp}, focusing on their ability to generalize across the two material systems and two mechanical properties. We then analyze the explainable parameters of GPs to gain insights into the underlying structure-property relationships in Section \ref{sec:gp-interpret}. Finally, we discuss the implications of our findings for data fusion and transfer learning in Section \ref{sec:results-discussion}, emphasizing the need for principled approaches that incorporate domain knowledge.

\subsection{Correlation Analyses}\label{sec:data-correlation}
To understand the relationships between process parameters and material properties, we first analyze the correlation between the two mechanical properties and volumetric energy density (VED) which is an engineered feature designed to study the compound effect of multiple process parameters on the LPBF-built part. VED is defined as:
\begin{equation}
    VED = \frac{p}{v \cdot h \cdot l}
\end{equation}

\begin{figure*}[!t]
    \centering
    \includegraphics[width=\textwidth]{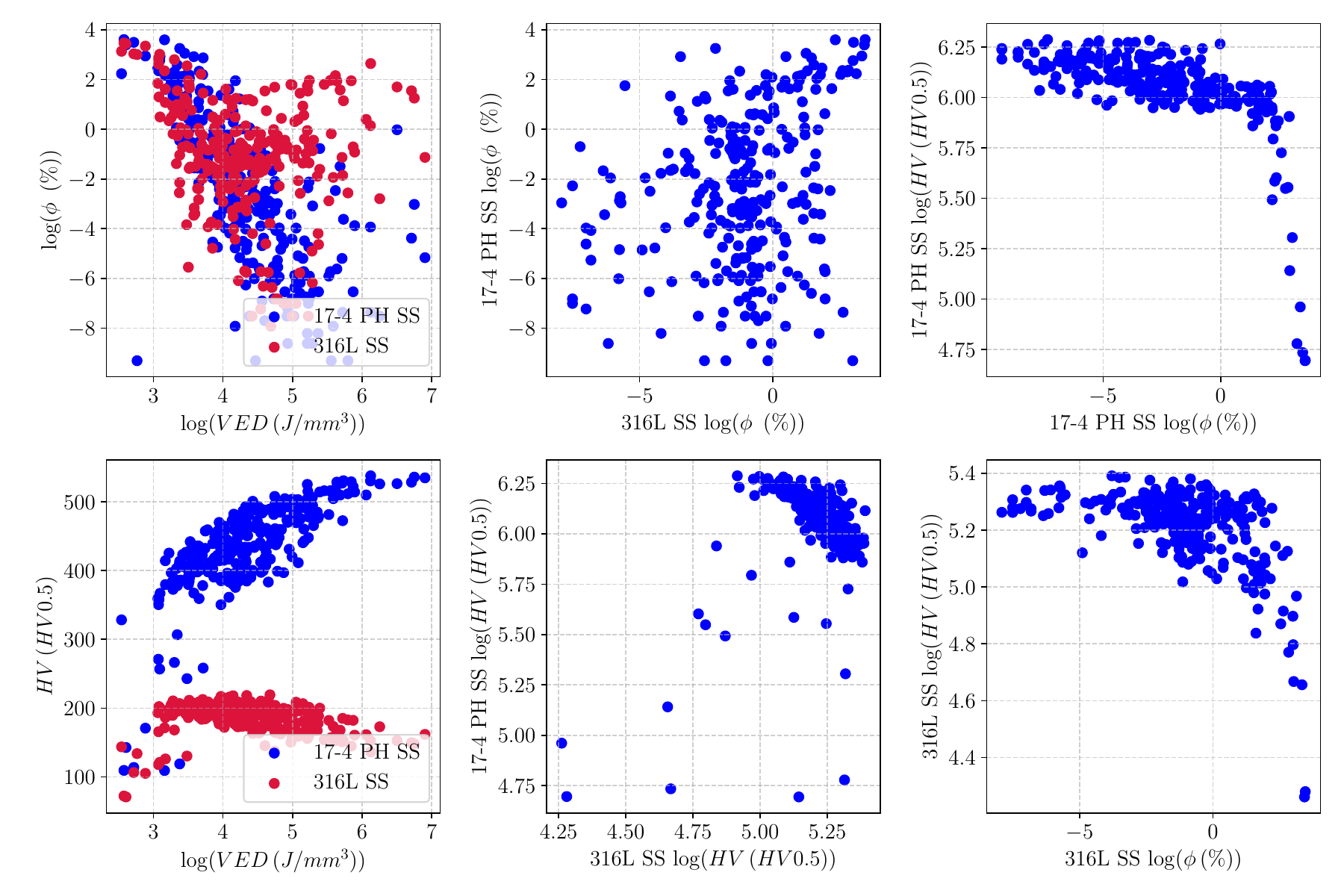}
    \caption{Left: correlations between VED (Log scale) and porosity (\porosity) and hardness (\hardness) properties. Center: cross-material correlations (log scale) for \alloy~and \secalloy~properties. Right: cross-property correlations (log scale) within \alloy~and \secalloy.}
    \label{fig:material-property-correlations}
\end{figure*}

Figure \ref{fig:material-property-correlations} shows scatter plots of VED against \porosity~and \hardness, with data points color-coded based on the material type (\alloy~and \secalloy). These plots reveal the stochasticity of the data, as well as the presence of underlying trends that suggest a correlation between VED and the properties. As demonstrated in prior works \cite{Ion1992DiagramsProcessing,Thomas2016NormalisedAlloys}, this indicates that process-property models can be developed to predict material behavior.

While VED is a very useful engineered featured that simplifies the visualization of the complex effects that process parameters have on a property of interest, it prevents designers to probe the impact of individual process parameters on the properties. To address this issue, prior work has leveraged data-driven techniques \cite{amiri2024unveiling,Huang2021HighOptimization,Agrawal2020High-throughputSteel} where multiple samples are first built under various combinations of process parameters and then their properties (e.g., hardness and porosity) are obtained via appropriate characterization methods. Afterwards, the resulting dataset is used to build ML models that aim to learn the relation between process parameters and the characterized properties. We follow a similar approach in this paper.

\begin{table}[!b]
\caption{Pearson correlation coefficients for \alloy~and \secalloy.}
\centering
\begin{tabular}{cccccc}
\cline{3-6}
\multicolumn{2}{c}{} & \multicolumn{2}{c}{\alloy} & \multicolumn{2}{c}{\secalloy} \\ \cline{3-6}
& & \porosity & \hardness & \porosity & \hardness \\ \hline
\multicolumn{1}{c}{\multirow{2}{*}{\alloy}} & \porosity & 1.00 & -0.83 & 0.66 & -0.31    \\
\multicolumn{1}{c}{} & \hardness &  & 1.00  & -0.53 &  0.02   \\ \hline
\multicolumn{1}{c}{\multirow{2}{*}{\secalloy}} & \porosity &  &  & 1.00 &  -0.71   \\
\multicolumn{1}{c}{} & \hardness &  &  &  & 1.00    \\
\hline
\end{tabular}
\label{tab: pearson}
\end{table}
To further investigate the transferability of information across material systems and properties, we analyze the correlations between \porosity~and \hardness~for each material. The corresponding scatter plots are provided in Figure \ref{fig:material-property-correlations}, which indicates that \hardness~and \porosity~are inversely correlated in each material system: in both \alloy~and \secalloy, \hardness~gradually decreases as \porosity~increases up to a certain value after which the drop in the \hardness~is substantial. This is an expected behavior since highly porous samples have small hardness and strength. 

Interesting trends are observed across the two materials: while there is a weak positive correlation between porosity values, the measured \hardness~values for \alloy~and \secalloy~seem to be negatively correlated. That is, on average, a particular change in process parameters increases the \hardness~of \alloy~while decreasing that of \secalloy. 

\begin{table}[!t]
\caption{Spearman correlation coefficients for \alloy~and \secalloy.}
\centering
\begin{tabular}{cccccc}
\cline{3-6}
\multicolumn{2}{c}{} & \multicolumn{2}{c}{\alloy} & \multicolumn{2}{c}{\secalloy} \\ \cline{3-6}
& & \porosity & \hardness & \porosity & \hardness \\ \hline
\multicolumn{1}{c}{\multirow{2}{*}{\alloy}} & \porosity & 1.00 & -0.76 & 0.21 & -0.35    \\
\multicolumn{1}{c}{} & \hardness &  & 1.00  & 0.05 &  -0.58   \\ \hline
\multicolumn{1}{c}{\multirow{2}{*}{\secalloy}} & \porosity &  &  & 1.00 &  -0.47   \\
\multicolumn{1}{c}{} & \hardness &  &  &  & 1.00    \\
\hline
\end{tabular}
\label{tab: sperman}
\end{table}

To quantify these trends, Table \ref{tab: pearson} and Table \ref{tab: sperman} report the Pearson and Spearman rank correlations, respectively. These results confirm a strong negative cross-property correlation within each material system—e.g., Pearson correlation between \porosity~and \hardness~for \alloy~is -0.83—, reinforcing that increased porosity leads to lower \hardness~in both alloys. However, cross-material correlations for individual properties are notably weaker than within-material trends—e.g., Spearman correlation between \alloy~and \secalloy~for \porosity~is 0.21.

\subsection{GP-based Process-Property Models}\label{sec:models-comp}

We train SOGPs—predicting \porosity~or \hardness~separately—and MTGPs—jointly predicting \porosity~and \hardness—with and without data fusion. As shown in Table \ref{tab:rmse-table}, the performance of these six models is quantified via RMSE averaged across 5-fold cross-validation. 

\begin{table}[!b]
    \caption{RMSE comparison for each GP configuration using 5-fold cross-validation.}
    \centering
    \begin{tabular}{ccccc}
    \hline
        Model & Train & Test & \textbf{\porosity $(\%)$} & \textbf{\hardness~$(HV0.5)$}  \\ \hline
        \multirow{3}{*}{SOGP} & \alloy & \alloy & 2.18 & 17.82 \\
                                       & \secalloy & \secalloy & 2.15 & 8.41  \\
                                       & \multirow{2}{*}{Fused Data} & \alloy & 2.38 & 18.01  \\
                                                                  &  & \secalloy & 1.84 & 8.6  \\ \hline
        \multirow{3}{*}{MTGP}  & \alloy & \alloy & 2.35 & 17.89 \\
                                       & \secalloy & \secalloy & 1.94 & 8.21 \\
                                       & \multirow{2}{*}{Fused Data} & \alloy & 6.14 & 18.35 \\
                                                                   & & \secalloy & 2.12 & 8.92 \\ \hline
    \end{tabular}
    \label{tab:rmse-table}
\end{table}

First, we analyze the impact of data fusion by comparing models trained on either one or two materials data. Using fused data slightly increases RMSE of SOGPs for both properties in \alloy: from $2.18\,\%$ to $2.38\,\%$ for \porosity~and from $17.82\,HV0.5$ to $18.01\,HV0.5$ for hardness. In the case of \secalloy, SOGP with fused data slightly improves \porosity~prediction but worsens \hardness~prediction. For MTGPs, data fusion's effects are more pronounced, particularly for \alloy. The \porosity~RMSE jumps from $2.35\,\%$ to $6.14\,\%$, and \hardness~RMSE also increases from $17.89\,HV0.5$ to $18.35\,HV0.5$. Similarly, data fusion is ineffective for \secalloy~as a single-material MTGP predicts both \porosity~and \hardness~better than a bi-material MTGP. 
These results suggest that data fusion does not consistently enhance predictive performance in our application since the GPs struggle to generalize across materials when trained on the fused dataset.

Next, we compare MTGP to SOGP to assess the effect of incorporating cross-property modeling. In the case of \alloy, SOGP achieves slightly lower RMSEs ($2.18\,\%$ and $17.82\,HV0.5$) compared to MTGP trained on \alloy~($2.35\,\%$ and $17.89\,HV0.5$), while for \secalloy~we observe the opposite and MTGP obtains slightly lower RMSEs ($1.94\,\%$ and $8.21\,HV0.5$) than the ones obtained in SOGP ($2.15\,\%$ and $8.41\,HV0.5$). These results indicate that MTGP does not consistently outperform SOGP and vice versa; suggesting that the correlations between \porosity~and \hardness~are either very small, or cannot be learned via our MTGPs. The latter case can be due to a number of factors such as simplicity of our covariance function or lack of data. 

Finally, we compare the SOGP model trained on fused data with the MTGP model trained on fused data to assess the impact of both cross-material and cross-property correlations. We observe that for both properties (\porosity~ and \hardness) evaluated on each of the materials MTGP model exhibits worse RSMEs than the SOGPs, confirming that MTGPs fail to leverage the correlations hidden in the data.

\begin{table*}[!t]
\caption{Learned lengthscale parameters for input features across SOGP and MTGP models.}
\centering
\begin{tabular}{lccccccc}
\hline
Method & Material & $p$ & $v$ & $l$ & $h$ & $sr$ & $s$ \\ \hline
\multirow{3}{*}{SOGP \porosity} 
    & \alloy   & -1.06 & -0.84 & -1.14 & -1.92 & -2.30 & -    \\
    & \secalloy     & -0.42 & -0.24 & -0.66 & -0.44 & -2.17 & -    \\
    & Fused Data & -0.83 & -0.71 & -1.10 & -1.55 & -2.31 & 0.93 \\ \hline
\multirow{3}{*}{SOGP \hardness}
    & \alloy   & -0.90 & -1.05 & -1.95 & -2.32 & -2.32 & -    \\
    & \secalloy     & -0.83 & -0.50 & -2.22 & -2.18 & -2.15 & -    \\
    & Fused Data & -0.98 & -1.19 & -2.27 & -2.36 & -2.36 & 1.30 \\ \hline
\multirow{3}{*}{MTGP} 
    & \alloy   & -0.98 & -1.08 & -1.23 & -2.32 & -2.31 & -    \\
    & \secalloy     & -0.76 & -0.56 & -1.57 & -1.65 & -2.15 & -    \\
    & Fused Data & -1.20 & -1.14 & -1.97 & -2.34 & -2.36 & 1.37 \\ \hline
\end{tabular}
\label{tab: lengthscale_params}
\end{table*}

\begin{figure*}[!b]
    \centering
    \begin{subfigure}[b]{0.48\textwidth}
        \includegraphics[width=\textwidth]{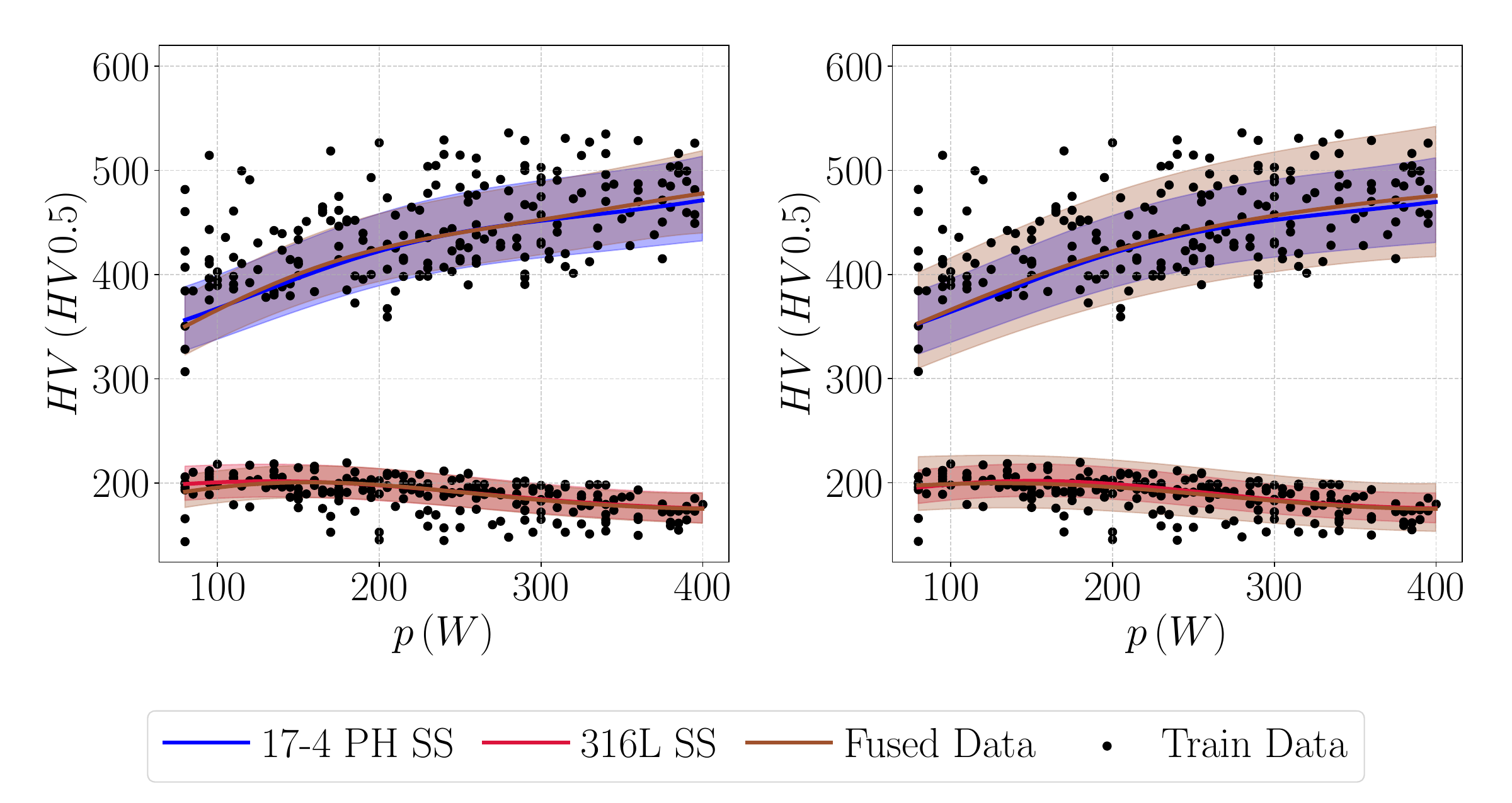}
        \caption{Hardness ($HV$) predictive posterior distribution as a function of laser power ($p$). Left: SOGP. Right: MTGP.}
        \label{fig:power-hardness-all}
    \end{subfigure}
    \hspace{0.01\textwidth}
    \begin{subfigure}[b]{0.48\textwidth}
        \includegraphics[width=\textwidth]{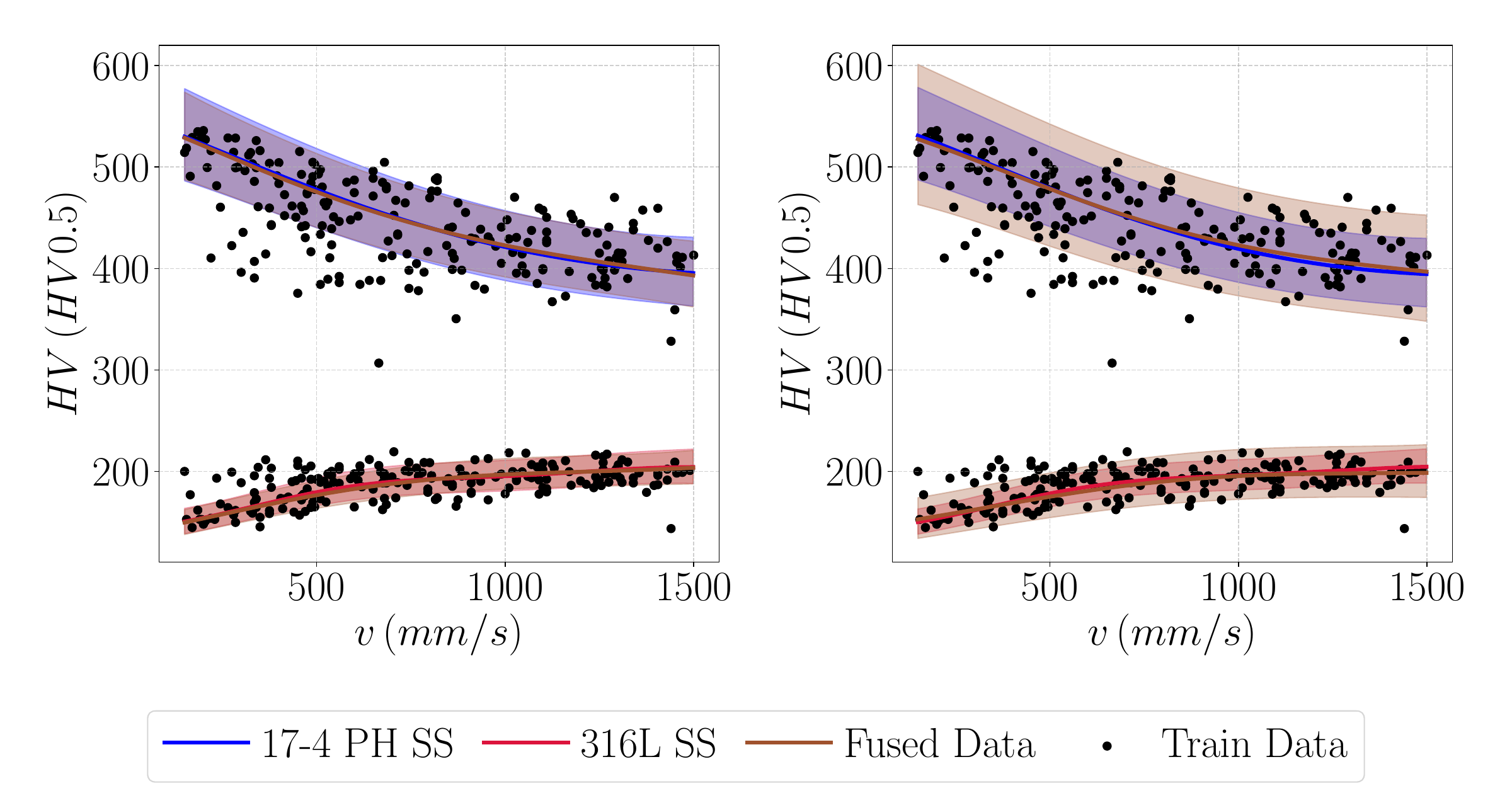}
        \caption{Hardness ($HV$) predictive posterior distribution as a function of laser scan speed ($v$). Left: SOGP. Right: MTGP.}
        \label{fig:speed-hardness-all}
    \end{subfigure}
    \caption{Marginal feature analysis of Hardness (\hardness) predictive posterior distributions for SOGP and MTGP models.}
    \label{fig:pores}
\end{figure*}

Overall, our results indicate that data fusion does not significantly improve predictive accuracy, and incorporating cross-property correlations in the MTGP through a shared kernel structure does not lead to better performance than the SOGP models. That is, the two datasets entail sufficiently distinct relationships between process parameters and properties that a bi-material model does not learn a robust shared pattern that outperforms material-specific models.  The complex relationship between \porosity~and \hardness~is influenced by microstructural differences between \alloy~ and \secalloy—such as variations in grain size, thermal history, and phase composition—limiting the effectiveness of a generalized modeling approach.

The findings align with the observations from Section \ref{sec:data-correlation}, where Pearson and Spearman correlations indicate stronger relationships within single-material systems compared to cross-material systems and cross-property relationships. Simply combining datasets from different materials does not necessarily lead to improved predictions in LPBF, even when the materials share similar processing conditions.

To further confirm our results, we also explore the use of a richer set of features extracted from the porosity images (e.g., pore radius, pore location), as outlined in Section \ref{sec: AM_design}, to potentially improve predictive accuracy. However, the results obtained from this richer feature set are comparable to those achieved using the average porosity, indicating that considering additional responses (which increase complexity too) do not provide a significant performance boost. For brevity, we do not report these results as they would be very similar to the above descriptions.

\begin{figure*}[!b]
    \centering
    \begin{subfigure}[b]{0.48\textwidth}
        \includegraphics[width=\textwidth]{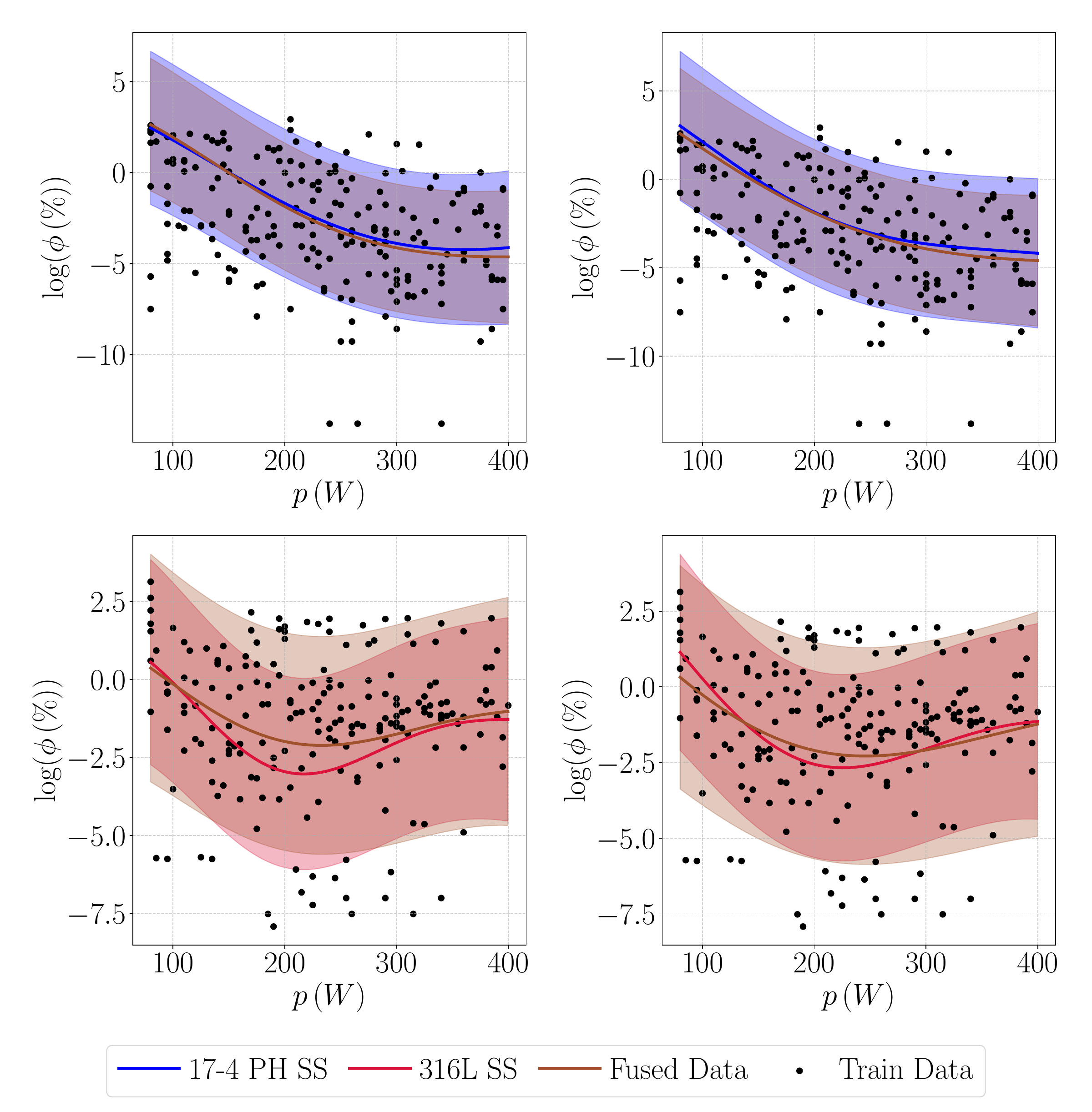}
        \caption{Porosity ($\phi$) predictive posterior distribution as a function of laser power ($p$). Left: SOGP. Right: MTGP.}
        \label{fig:power-porosity-all}
    \end{subfigure}
    \hspace{0.01\textwidth}
    \begin{subfigure}[b]{0.48\textwidth}
        \includegraphics[width=\textwidth]{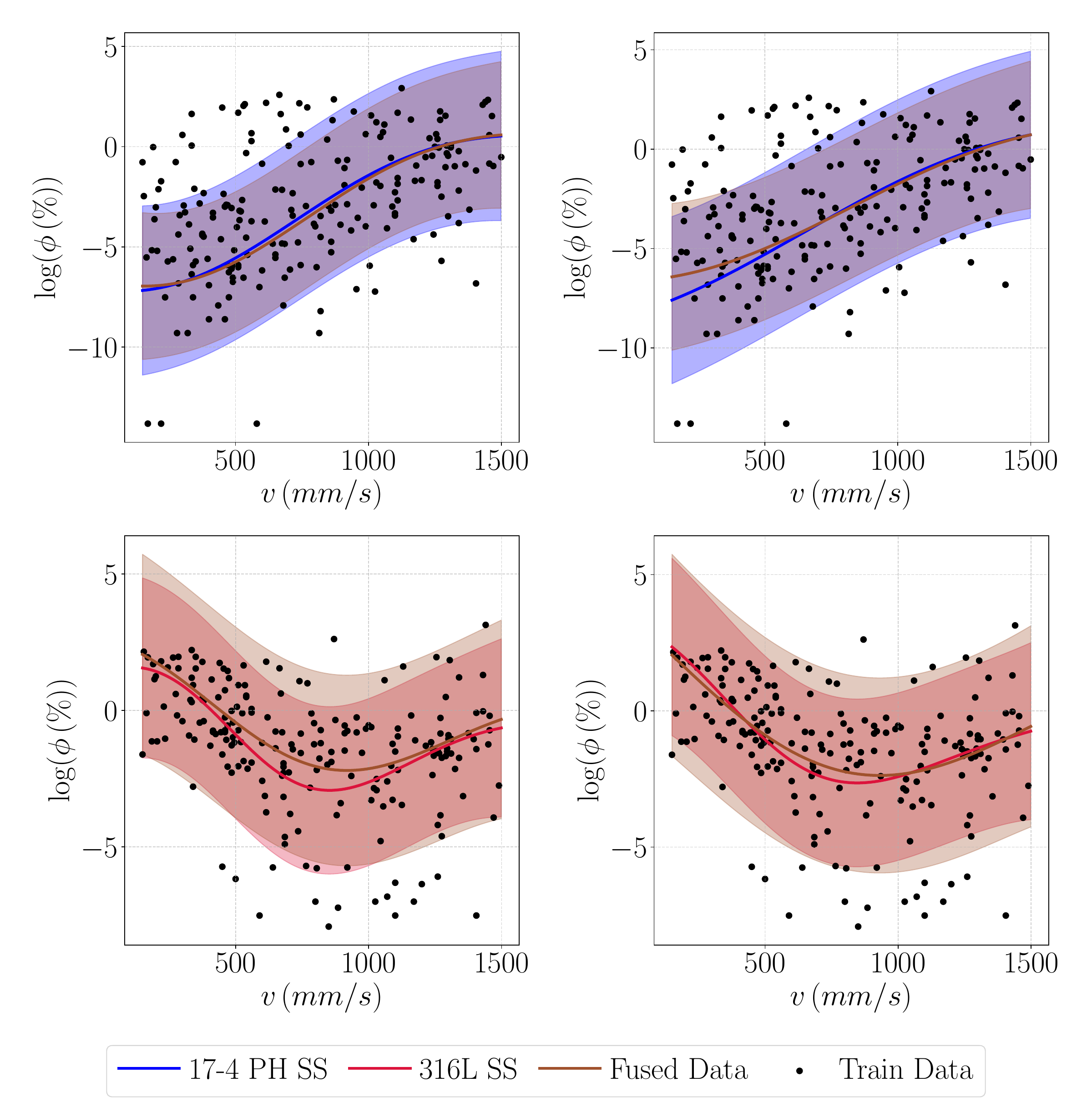}
        \caption{Porosity ($\phi$) predictive posterior distribution as a function of laser scan speed ($v$). Left: SOGP. Right: MTGP.}
        \label{fig:speed-porosity-all}
    \end{subfigure}
    \caption{Marginal feature analysis of Porosity (\porosity) predictive posterior distributions for SOGP and MTGP models.}
    \label{fig:pores}
\end{figure*}

\subsection{Interpretation of Model Parameters}\label{sec:gp-interpret}

Results from Section \ref{sec:models-comp} do not show significant performance improvement from fusing the two material systems. In this Section, we investigate the hyperparameters of our GPs to assess whether they point to the same conclusion. 
Specifically, we analyze the lengthscale hyperparameters in Equation \ref{eq: rbf-kernel}, i.e., $\omegab$, to understand the underlying feature relative importance for process-property modeling. Table \ref{tab: lengthscale_params} shows the lengthscale values for the SOGPs  and MTGPs. 

We first analyze the impact of data fusion by comparing the lengthscale parameters for each bi-material model with its corresponding single-material version. The lengthscale parameter corresponding to the categorical source indicator variable (i.e., $s$) in the bi-material models is significantly larger than the largest lengthscales corresponding to the process parameters. For example, in the SOGP \hardness~model the lengthscale for $s$ is $1.30$ while for $p$ is $-0.98$. Such a difference indicates that the two materials are effectively independent since for the covariance function in Equation \ref{eq: rbf-kernel} a large $\omega_i$ indicates that the correlations die out very quickly as feature $i$ varies. 

Next, we asses the effect of incorporating cross-property relationships into the model by comparing MTGPs to SOGPs. While there are some differences, most of the lengthscale parameters have similar values. For example, the lengthscale for $h$ is $-0.44$ in the SOGP \porosity~\secalloy~model and $-1.65$ in the MTGP \secalloy~model.
Unlike $h$, the lengthscale for $v$ is $-1.05$ in SOGP \hardness~\alloy~and $-1.08$ in MTGP. We also observe that excluding $s$ in the bi-material models, $p$ and $v$ are the most important process parameters in terms of affecting the responses while $h$ and $sr$ are the least important ones. This is true across all SOGP and MTGP models, although the exact order is not always preserved—e.g., for \alloy~in the SOGP \porosity~model $v$ ($-0.84$) has more effect on the properties than $p$ ($-1.06$) while in the MTGP model $p$ ($-0.98$) is slightly more important than $v$ ($-1.08$).

In MGTP we also analyze the off-diagonal term of $\boldsymbol{C}_T$, which captures the corelation between \porosity~and \hardness. For each of the three MGTP models (\alloy, \secalloy~and Fused Data), the off-diagonal term is negative and it indicates a negative correlation between the two properties. This is aligned with the Pearson and Spearman coefficients analyzed in Section \ref{sec:data-correlation}. Nonetheless, this cross-property correlation does not improve the performance of the model.


Building on the above analysis, we select the two most important process parameters—$p$ and $v$— and visualize their effects on the properties while fixing all the other parameters at their median values. We provide the plots for three conditions: using only \alloy, using only \secalloy, and combining both material systems through data fusion.
The results are shown in Figures \ref{fig:power-hardness-all}, \ref{fig:speed-hardness-all}, \ref{fig:power-porosity-all}, \ref{fig:speed-porosity-all} where we also show the training samples (note that the training data involve simultaneous variations in all parameters and should not be directly compared to the illustrated models' predictions).

We see in Figure \ref{fig:power-hardness-all} that the two materials exhibit different trends: as $p$ increases \alloy~becomes harder while that hardness of\secalloy~slightly decreases. Looking at Figure \ref{fig:speed-porosity-all} we observe that the porosity of \alloy~and \secalloy~depend on $v$ quite differently. 
Comparing the plots based on SOGPs and MTGPs, we can conclude that the models' predictions are quite similar, with the MTGPs sometimes providing slightly wider predictions intervals. 
Consistent with the previous analyses, we observe in these figures that combining the two materials data negligibly affect the predictions. 


\subsection{Implications for Data Fusion and Transfer Learning}\label{sec:results-discussion}

The results of this work offer interesting insights for the development of predictive models in additive manufacturing. Specifically, they challenge the prevailing assumption that materials with similar processing conditions inherently benefit from data fusion. Instead, our findings suggest that material system compatibility, driven by shared underlying physical mechanisms, plays a far more crucial role in determining the success of data fusion. This implies that simply increasing dataset size by combining data from different materials does not necessarily lead to improved predictive accuracy, as our results demonstrate.

These observations are particularly relevant for applications in data-limited domains like LPBF, where the assumption of universal transferability across material systems may not hold. Instead of indiscriminate data fusion, our results advocate for more structured approaches to transfer learning—ones that carefully account for intrinsic differences between materials. By considering the unique characteristics of each material system, these methods could better capture the complex process-property relationships that govern material behavior.

%% file: 15_Conclusion.tex
\section{Conclusions} \label{sec: conclusion}

Our study examines the feasibility of data transferability in LPBF as a preliminary step toward developing foundation process-property models. Leveraging GPs, we assess the potential benefits of fusing data from \alloy~and \secalloy~material systems. The results demonstrate that data fusion, in this context, does not improve predictive accuracy even with a shared input feature space, suggesting that material system compatibility is a critical factor beyond shared processing conditions. This finding implies that fusing data within related material systems does not guarantee enhanced model performance and it suggests that scaling data without careful consideration of feature representation may not yield to robust and generalizable foundation process-property models in LPBF.

While this study provides valuable insights, it has some limitations. First, the dataset size, though reasonable for additive manufacturing, is relatively small compared to other machine learning domains. Due to the small dataset size, we used GPs with parsimonious kernels that avoid overfitting. Leveraging more flexible kernels or ML models with larger datasets should be considered in the future studies. 
Second, the analysis is limited to two materials and the generalizability of our findings across a broader range of materials must be explored.

\section{Acknowledgments}
 We appreciate the support from UC National Laboratory Fees Research Program of the University of California (Grant Number $L22CR4520$). Oriol Vendrell-Gallart also acknowledges the Balsells Fellowship.